# A Convex Formulation for Learning Task Relationships in Multi-Task Learning

**Yu Zhang & Dit-Yan Yeung**
Department of Computer Science and Engineering, Hong Kong University of Science and Technology
Clear Water Bay, Kowloon, Hong Kong, China
{zhangyu,dyyeung}@cse.ust.hk

## Abstract

Multi-task learning is a learning paradigm which seeks to improve the generalization performance of a learning task with the help of some other related tasks. In this paper, we propose a regularization formulation for learning the relationships between tasks in multi-task learning. This formulation can be viewed as a novel generalization of the regularization framework for single-task learning. Besides modeling positive task correlation, our method, called *multi-task relationship learning* (MTRL), can also describe negative task correlation and identify outlier tasks based on the same underlying principle. Under this regularization framework, the objective function of MTRL is convex. For efficiency, we use an alternating method to learn the optimal model parameters for each task as well as the relationships between tasks. We study MTRL in the symmetric multi-task learning setting and then generalize it to the asymmetric setting as well. We also study the relationships between MTRL and some existing multi-task learning methods. Experiments conducted on a toy problem as well as several benchmark data sets demonstrate the effectiveness of MTRL.

## 1 Introduction

Multi-task learning [9, 4, 24] is a learning paradigm which seeks to improve the generalization performance of a learning task with the help of some other related tasks. This learning paradigm has been inspired by human learning activities in that people often apply the knowledge gained from previous learning tasks to help learn a new task. For example, a baby first learns to recognize human faces and later uses this knowledge to help it learn to recognize other objects. Multi-task learning can be formulated under two different settings: *symmetric* and *asymmetric* [26]. While symmetric multi-task learning seeks to improve the performance of all tasks simultaneously, the objective of asymmetric multi-task learning is to improve the performance of some target task using information from the source tasks, typically after the source tasks have been learned using some symmetric multi-task learning method. In this sense, asymmetric multi-task learning is related to *transfer learning* [22], but the major difference is that the source tasks are still learned simultaneously in asymmetric multi-task learning but they are learned independently in transfer learning.

Major advances have been made in multi-task learning over the past decade, although some preliminary ideas actually date back to much earlier work in psychology and cognitive science. Multi-layered feedforward neural networks provide one of the earliest models for multi-task learning. In a neural network, the hidden layer represents the common features for data points from all tasks and each unit in the output layer usually corresponds to the output of one task. Similar to neural networks, multi-task feature learning [1, 20] also learns common features for all tasks but under the regularization framework. Unlike these methods, the regularized multi-task support vector machine (SVM) [12] enforces the SVM parameters for all tasks to be close to each other. Another widely studied approach for multi-task learning is the task clustering approach [25, 3, 26]. Its main idea is to group the tasks into several clusters and then learn similar data features or model parameters for the tasks within each cluster. An advantage of this approach is its robustness against outlier tasks because they reside in separate clusters that do not affect other tasks. Moreover, some Bayesian models have been proposed for multi-task learning [7, 15, 27]. Most of the above methods focus on *symmetric* multi-task learning, but there also exist some previous works that study *asymmetric* multi-task learning [23, 19, 10].

Since multi-task learning seeks to improve the performance of a task with the help of other related tasks, a central issue is to characterize the relationships between tasks accurately. Given the training data in multiple tasks, there are two important aspects that distinguish between differ-ent methods for characterizing the task relationships. The first aspect is on *what* task relationships can be represented by a method. Generally speaking there are three types of pairwise task relationships: positive task correlation, negative task correlation, and task unrelatedness (corresponding to outlier tasks). Positive task correlation is very useful to characterize the task relationship since similar tasks are likely to have similar model parameters. For negative task correlation, since the model parameters of two tasks with negative correlation are more likely to be dissimilar, knowing that two tasks are negatively correlated can help to reduce the search space of the model parameters. As for task unrelatedness, identifying outlier tasks can prevent them from impairing the performance of other tasks since outlier tasks are unrelated to other tasks. The second aspect is on *how* to obtain the relationships, either from the model assumption or automatically learned from data. Obviously learning the task relationships from data automatically is the more favorable option because the model assumption adopted may be incorrect and, even worse, it is not easy to verify the correctness of the assumption from data.

Multi-layered feedforward neural networks and multi-task feature learning assume that all tasks share the same representation without actually learning the task relationships from data automatically. Moreover, they do not consider negative task correlation and are not robust against outlier tasks. The regularization methods in [12, 11, 17] assume that the task relationships are given and then utilize this prior knowledge to learn the model parameters. However, they just utilize positive task correlation and task unrelatedness but not negative task correlation. The task clustering methods in [25, 3, 26, 16] may be viewed as a way to learn the task relationships from data. Similar tasks will be grouped into the same task cluster and outlier tasks will be grouped separately, making these methods more robust against outlier tasks. However, they are *local* methods in the sense that only similar tasks within the same task cluster can interact to help each other, thus ignoring negative task correlation which may exist between tasks residing in different clusters. On the other hand, a powerful multi-task learning method based on Gaussian process (GP) [7] provides a *global* approach to model and learn task relationships in the form of a task covariance matrix. A task covariance matrix can model all three types of task relationships: positive task correlation, negative task correlation, and task unrelatedness. However, although this method provides a powerful way to model task relationships, learning of the task covariance matrix gives rise to a non-convex optimization problem which is sensitive to parameter initialization. When the number of tasks is large, the authors proposed to use low-rank approximation [7] which will then weaken the expressive power of the task covariance matrix. Moreover, since the method is based on GP, scaling it to large data sets poses a serious computational challenge.

Our goal in this paper is to inherit the advantages of [7] while overcoming its disadvantages. Specifically, we propose a method, called *multi-task relationship learning* (MTRL), which also models the relationships between tasks in a nonparametric manner as a task covariance matrix. Based on a regularization framework, we obtain a convex objective function which allows us to learn the model parameters and the task relationships simultaneously. MTRL can be viewed as a generalization of the regularization framework for single-task learning to the multi-task setting. For efficiency, we use an alternating optimization method in which each subproblem is a convex problem. We study MTRL in the symmetric multi-task learning setting and then generalize it to the asymmetric setting as well. We also study the relationships between MTRL and some existing multi-task learning methods, showing that these methods can be viewed as special cases of MTRL.

## 2 Multi-Task Relationship Learning

Suppose we are given $m$ learning tasks $\{T_i\}_{i=1}^m$. For the $i$th task $T_i$, the training set $\mathcal{D}_i$ consists of $n_i$ data points $(\mathbf{x}_j^i, y_j^i)$, $j = 1, \ldots, n_i$, with $\mathbf{x}_j^i \in \mathbb{R}^d$ and its corresponding output $y_j^i \in \mathbb{R}$ if it is a regression problem and $y_j^i \in \{-1, 1\}$ if it is a binary classification problem. The linear function for $T_i$ is defined as $f_i(\mathbf{x}) = \mathbf{w}_i^T \mathbf{x} + b_i$.

### 2.1 Probabilistic Framework

The likelihood for $y_j^i$ given $\mathbf{x}_j^i$, $\mathbf{w}_i$, $b_i$ and $\varepsilon_i$ is:

$$y_j^i \,|\, \mathbf{x}_j^i, \mathbf{w}_i, b_i, \varepsilon_i \sim \mathcal{N}(\mathbf{w}_i^T \mathbf{x}_j^i + b_i, \varepsilon_i^2), \tag{1}$$

where $\mathcal{N}(\mathbf{m}, \boldsymbol{\Sigma})$ denotes the multivariate (or univariate) normal distribution with mean $\mathbf{m}$ and covariance matrix (or variance) $\boldsymbol{\Sigma}$.

The prior on $\mathbf{W} = (\mathbf{w}_1, \ldots, \mathbf{w}_m)$ is defined as

$$\mathbf{W}|\epsilon_i \sim \left(\prod_{i=1}^m \mathcal{N}(\mathbf{w}_i|\mathbf{0}_d, \epsilon_i^2 \mathbf{I}_d)\right) q(\mathbf{W}), \tag{2}$$

where $\mathbf{I}_d$ is the $d \times d$ identity matrix. The first term of the prior on $\mathbf{W}$ is to penalize the complexity of each column of $\mathbf{W}$ separately and the second term is to model the structure of $\mathbf{W}$. Since $\mathbf{W}$ is a matrix variable, it is natural to use a matrix-variate distribution [14] to model it. Here we use the matrix-variate normal distribution for $q(\mathbf{W})$. More specifically,

$$q(\mathbf{W}) = \mathcal{MN}_{d \times m}(\mathbf{W} \,|\, \mathbf{0}_{d \times m}, \mathbf{I}_d \otimes \boldsymbol{\Omega}) \tag{3}$$

where $\mathcal{MN}_{d \times m}(\mathbf{M}, \mathbf{A} \otimes \mathbf{B})$ denotes a matrix-variate normal distribution[1] with mean $\mathbf{M} \in \mathbb{R}^{d \times m}$, row covariance matrix $\mathbf{A} \in \mathbb{R}^{d \times d}$ and column covariance matrix $\mathbf{B} \in \mathbb{R}^{m \times m}$. The row covariance matrix $\mathbf{I}_d$ models the relationships between features and the column covariance matrix $\mathbf{\Omega}$ models the relationships between different $\mathbf{w}_i$'s. In other words, $\mathbf{\Omega}$ models the relationships between tasks.

---

[1] The probability density function is defined as $p(\mathbf{X} \,|\, \mathbf{M}, \mathbf{A}, \mathbf{B}) = \frac{\exp\left(-\frac{1}{2}\mathrm{tr}\left(\mathbf{A}^{-1}(\mathbf{X}-\mathbf{M})\mathbf{B}^{-1}(\mathbf{X}-\mathbf{M})^T\right)\right)}{(2\pi)^{md/2}|\mathbf{A}|^{m/2}|\mathbf{B}|^{d/2}}$.

When there is only one task and $\mathbf{\Omega}$ is given as a positive scalar, our model will degenerate to the probabilistic model for regularized least-squares regression and least-squares SVM [13]. So our probabilistic model can be viewed as a generalization of the probabilistic model for single-task learning. However, unlike single-task learning, $\mathbf{\Omega}$ cannot be given *a priori* for most multi-task learning applications and so we seek to estimate it from data automatically.

It follows that the posterior distribution for $\mathbf{W}$, which is proportional to the product of the prior and the likelihood function [6], is given by:

$$p(\mathbf{W} \mid \mathbf{X}, \mathbf{y}, \mathbf{b}, \boldsymbol{\varepsilon}, \boldsymbol{\epsilon}, \mathbf{\Omega}) \propto p(\mathbf{y} \mid \mathbf{X}, \mathbf{W}, \mathbf{b}, \boldsymbol{\varepsilon}) p(\mathbf{W} \mid \boldsymbol{\epsilon}, \mathbf{\Omega}), \quad (4)$$

where $\mathbf{X}$ denotes the data matrix of all data points in all tasks, $\mathbf{y} = (y_1^1, \ldots, y_{n_1}^1, \ldots, y_1^m, \ldots, y_{n_m}^m)^T$, and $\mathbf{b} = (b_1, \ldots, b_m)^T$. Taking the negative logarithm of Eq. (4) and combining with Eqs. (1)–(3), we obtain the *maximum a posteriori* (MAP) estimation of $\mathbf{W}$ and the *maximum likelihood estimation* (MLE) of $\mathbf{b}$ and $\mathbf{\Omega}$ by minimizing

$$\min_{\mathbf{W},\mathbf{b},\mathbf{\Omega}} \sum_{i=1}^{m} \frac{1}{\varepsilon_i^2} \sum_{j=1}^{n_i} (y_j^i - \mathbf{w}_i^T \mathbf{x}_j^i - b_i)^2 + \sum_{i=1}^{m} \frac{1}{\epsilon_i^2} \mathbf{w}_i^T \mathbf{w}_i$$
$$+ \operatorname{tr}(\mathbf{W} \mathbf{\Omega}^{-1} \mathbf{W}^T) + d \ln |\mathbf{\Omega}|, \quad (5)$$

where $\operatorname{tr}(\cdot)$ denotes the trace of a square matrix and $|\cdot|$ denotes the determinant of a square matrix. For simplicity of discussion, we assume that $\varepsilon = \varepsilon_i$ and $\epsilon = \epsilon_i$, $\forall i = 1, \ldots, m$. The effect of the last term in problem (5) is to penalize the complexity of $\mathbf{\Omega}$. However, as we will see later, the first three terms in problem (5) are jointly convex with respect to all variables but the last term is concave since $-\ln |\mathbf{\Omega}|$ is a convex function with respect to $\mathbf{\Omega}$, according to [8]. Moreover, for kernel extension, we have no idea about $d$ which may even be infinite after feature mapping, making problem (5) difficult to optimize. So we replace the last term in problem (5) with a constraint $\operatorname{tr}(\mathbf{\Omega}) = 1$ to restrict its complexity, and problem (5) can be reformulated as

$$\min_{\mathbf{W},\mathbf{b},\mathbf{\Omega}} \quad \sum_{i=1}^{m} \sum_{j=1}^{n_i} (y_j^i - \mathbf{w}_i^T \mathbf{x}_j^i - b_i)^2 + \frac{\lambda_1}{2} \operatorname{tr}(\mathbf{W} \mathbf{W}^T)$$
$$+ \frac{\lambda_2}{2} \operatorname{tr}(\mathbf{W} \mathbf{\Omega}^{-1} \mathbf{W}^T)$$
$$\text{s.t.} \quad \mathbf{\Omega} \succeq 0$$
$$\operatorname{tr}(\mathbf{\Omega}) = 1, \quad (6)$$

where $\lambda_1 = \frac{2\varepsilon^2}{\epsilon^2}$ and $\lambda_2 = 2\varepsilon^2$ are regularization parameters, and $\mathbf{A} \succeq 0$ means that the matrix $\mathbf{A}$ is positive semidefinite. The first constraint in (6) holds due to the fact that $\mathbf{\Omega}$ is defined as a task covariance matrix. The first term in (6) measures the empirical loss on the training data, the second term penalizes the complexity of $\mathbf{W}$, and the third term measures the relationships between all tasks based on $\mathbf{W}$ and $\mathbf{\Omega}$.

To avoid the task imbalance problem in which one task has so many data points that it dominates the empirical loss, we modify problem (6) as

$$\min_{\mathbf{W},\mathbf{b},\mathbf{\Omega}} \quad \sum_{i=1}^{m} \frac{1}{n_i} \sum_{j=1}^{n_i} (y_j^i - \mathbf{w}_i^T \mathbf{x}_j^i - b_i)^2 + \frac{\lambda_1}{2} \operatorname{tr}(\mathbf{W} \mathbf{W}^T)$$
$$+ \frac{\lambda_2}{2} \operatorname{tr}(\mathbf{W} \mathbf{\Omega}^{-1} \mathbf{W}^T)$$
$$\text{s.t.} \quad \mathbf{\Omega} \succeq 0$$
$$\operatorname{tr}(\mathbf{\Omega}) = 1. \quad (7)$$

Note that (7) is a semi-definite programming (SDP) problem which is computationally demanding. In what follows, we will present an efficient algorithm for solving it.

### 2.2 Optimization Procedure

We first prove the convexity of problem (7) with respect to all variables.

**Theorem 1** *Problem (7) is convex with respect to $\mathbf{W}$, $\mathbf{b}$ and $\mathbf{\Omega}$.*

**Proof:**
It is easy to see that the first two terms in the objective function of problem (7) are convex with respect to all variables and the constraints in (7) are also convex. We rewrite the third term in the objective function as

$$\operatorname{tr}(\mathbf{W} \mathbf{\Omega}^{-1} \mathbf{W}^T) = \sum_t \mathbf{W}(t,:) \mathbf{\Omega}^{-1} \mathbf{W}(t,:)^T,$$

where $\mathbf{W}(t,:)$ denotes the $t$th row of $\mathbf{W}$. $\mathbf{W}(t,:) \mathbf{\Omega}^{-1} \mathbf{W}(t,:)^T$ is called a matrix fractional function in Example 3.4 (page 76) of [8] and it is proved to be a convex function with respect to $\mathbf{W}(t,:)$ and $\mathbf{\Omega}$ there when $\mathbf{\Omega}$ is a positive semidefinite matrix (which is satisfied by the first constraint of (7)). Since $\mathbf{W}(t,:)$ is a row of $\mathbf{W}$, $\mathbf{W}(t,:) \mathbf{\Omega}^{-1} \mathbf{W}(t,:)^T$ is also convex with respect to $\mathbf{W}$ and $\mathbf{\Omega}$. Because the summation operation can preserve convexity according to the analysis on page 79 of [8], $\operatorname{tr}(\mathbf{W} \mathbf{\Omega}^{-1} \mathbf{W}^T) = \sum_t \mathbf{W}(t,:) \mathbf{\Omega}^{-1} \mathbf{W}(t,:)^T$ is convex with respect to $\mathbf{W}$, $\mathbf{b}$ and $\mathbf{\Omega}$. So the objective function and the constraints in problem (7) are convex with respect to all variables and hence problem (7) is jointly convex. □

Even though the optimization problem (7) is convex with respect to $\mathbf{W}$, $\mathbf{b}$ and $\mathbf{\Omega}$ jointly, it is not easy to optimize the objective function with respect to all the variables simultaneously. Here we propose an alternating method to solve the problem more efficiently. Specifically, we first optimize the objective function with respect to $\mathbf{W}$ and $\mathbf{b}$ when $\mathbf{\Omega}$ is fixed, and then optimize it with respect to $\mathbf{\Omega}$ when $\mathbf{W}$ and $\mathbf{b}$ are fixed. This procedure is repeated until convergence. In what follows, we will present the two subproblems separately.

**Optimizing w.r.t. W and b when $\Omega$ is fixed**

When $\Omega$ is given and fixed, the optimization problem for finding $\mathbf{W}$ and $\mathbf{b}$ is an unconstrained convex optimization problem. The optimization problem can be stated as:

$$\min_{\mathbf{W},\mathbf{b}} \sum_{i=1}^{m} \frac{1}{n_i} \sum_{j=1}^{n_i} (y_j^i - \mathbf{w}_i^T \mathbf{x}_j^i - b_i)^2 + \frac{\lambda_1}{2} \text{tr}(\mathbf{W}\mathbf{W}^T)$$
$$+ \frac{\lambda_2}{2} \text{tr}(\mathbf{W}\Omega^{-1}\mathbf{W}^T). \quad (8)$$

To facilitate a kernel extension to be given later for the general nonlinear case, we reformulate the optimization problem into a dual form by first expressing problem (8) as a constrained optimization problem:

$$\min_{\mathbf{W},\mathbf{b},\{\varepsilon_j^i\}} \sum_{i=1}^{m} \frac{1}{n_i} \sum_{j=1}^{n_i} (\varepsilon_j^i)^2 + \frac{\lambda_1}{2} \text{tr}(\mathbf{W}\mathbf{W}^T) + \frac{\lambda_2}{2} \text{tr}(\mathbf{W}\Omega^{-1}\mathbf{W}^T)$$
$$\text{s.t.} \, y_j^i - (\mathbf{w}_i^T \mathbf{x}_j^i + b_i) = \varepsilon_j^i \quad \forall i,j. \quad (9)$$

The Lagrangian of problem (9) is given by

$$G = \sum_{i=1}^{m} \frac{1}{n_i} \sum_{j=1}^{n_i} (\varepsilon_j^i)^2 + \frac{\lambda_1}{2} \text{tr}(\mathbf{W}\mathbf{W}^T) + \frac{\lambda_2}{2} \text{tr}(\mathbf{W}\Omega^{-1}\mathbf{W}^T)$$
$$+ \sum_{i=1}^{m} \sum_{j=1}^{n_i} \alpha_j^i \left[ y_j^i - (\mathbf{w}_i^T \mathbf{x}_j^i + b_i) - \varepsilon_j^i \right]. \quad (10)$$

We calculate the gradients of $G$ with respect to $\mathbf{W}$, $b_i$ and $\varepsilon_j^i$ and set them to 0 to obtain

$$\frac{\partial G}{\partial \mathbf{W}} = \mathbf{W}(\lambda_1 \mathbf{I}_m + \lambda_2 \Omega^{-1}) - \sum_{i=1}^{m} \sum_{j=1}^{n_i} \alpha_j^i \mathbf{x}_j^i \mathbf{e}_i^T = 0$$

$$\Rightarrow \mathbf{W} = \sum_{i=1}^{m} \sum_{j=1}^{n_i} \alpha_j^i \mathbf{x}_j^i \mathbf{e}_i^T \Omega(\lambda_1 \Omega + \lambda_2 \mathbf{I}_m)^{-1}$$

$$\frac{\partial G}{\partial b_i} = -\sum_{j=1}^{n_i} \alpha_j^i = 0$$

$$\frac{\partial G}{\partial \varepsilon_j^i} = \frac{2}{n_i} \varepsilon_j^i - \alpha_j^i = 0,$$

where $\mathbf{e}_i$ is the $i$th column vector of $\mathbf{I}_m$. Combining the above equations, we obtain the following linear system:

$$\begin{pmatrix} \mathbf{K} + \frac{1}{2}\Lambda & \mathbf{M}_{12} \\ \mathbf{M}_{21} & \mathbf{0}_{m \times m} \end{pmatrix} \begin{pmatrix} \boldsymbol{\alpha} \\ \mathbf{b} \end{pmatrix} = \begin{pmatrix} \mathbf{y} \\ \mathbf{0}_{m \times 1} \end{pmatrix}, \quad (11)$$

where $k_{MT}(\mathbf{x}_{j_1}^{i_1}, \mathbf{x}_{j_2}^{i_2}) = \mathbf{e}_{i_1}^T \Omega(\lambda_1 \Omega + \lambda_2 \mathbf{I}_m)^{-1} \mathbf{e}_{i_2} (\mathbf{x}_{j_1}^{i_1})^T \mathbf{x}_{j_2}^{i_2}$ is the linear multi-task kernel, $\mathbf{K}$ is the kernel matrix defined on all data points for all tasks using the linear multi-task kernel, $\boldsymbol{\alpha} = (\alpha_1^1, \ldots, \alpha_{n_m}^m)^T$, $\mathbf{0}_{p \times q}$ is the $p \times q$ zero matrix or vector, $\Lambda$ is a diagonal matrix whose diagonal element is equal to $n_i$ if the corresponding data point belongs to the $i$th task, $N_i = \sum_{j=1}^{i} n_j$, and $\mathbf{M}_{12} = \mathbf{M}_{21}^T = (\mathbf{e}_{N_0+1}^{N_1}, \mathbf{e}_{N_1+1}^{N_2}, \ldots, \mathbf{e}_{N_{m-1}+1}^{N_m})$ where $\mathbf{e}_q^p$ is a zero vector with only the elements whose indices are in $[q, p]$ being equal to 1.

When the total number of data points for all tasks is very large, the computational cost required to solve the linear system (11) directly will be very high. In this situation, we can use another optimization method to solve it. It is easy to show that the dual form of problem (9) can be formulated as:

$$\min_{\boldsymbol{\alpha}} \quad h(\boldsymbol{\alpha}) = \frac{1}{2} \boldsymbol{\alpha}^T \tilde{\mathbf{K}} \boldsymbol{\alpha} - \sum_{i,j} \alpha_j^i y_j^i$$
$$\text{s.t.} \quad \sum_j \alpha_j^i = 0 \quad \forall i, \quad (12)$$

where $\tilde{\mathbf{K}} = \mathbf{K} + \frac{1}{2}\Lambda$. Note that it is similar to the dual form of least-squares SVM [13] except the difference that there is only one constraint in least-squares SVM but here there are $m$ constraints with each constraint corresponding to one task. Here we use an SMO algorithm similar to that for least-squares SVM [18].

**Optimizing w.r.t. $\Omega$ when W and b are fixed**

When $\mathbf{W}$ and $\mathbf{b}$ are fixed, the optimization problem for finding $\Omega$ becomes

$$\min_{\Omega} \quad \text{tr}(\Omega^{-1}\mathbf{W}^T\mathbf{W})$$
$$\text{s.t.} \quad \Omega \succeq 0$$
$$\text{tr}(\Omega) = 1. \quad (13)$$

Then we have

$$\text{tr}(\Omega^{-1}\mathbf{A}) = \text{tr}(\Omega^{-1}\mathbf{A})\text{tr}(\Omega)$$
$$= \text{tr}((\Omega^{-\frac{1}{2}}\mathbf{A}^{\frac{1}{2}})(\mathbf{A}^{\frac{1}{2}}\Omega^{-\frac{1}{2}}))\text{tr}(\Omega^{\frac{1}{2}}\Omega^{\frac{1}{2}})$$
$$\geq (\text{tr}(\Omega^{-\frac{1}{2}}\mathbf{A}^{\frac{1}{2}}\Omega^{\frac{1}{2}}))^2 = (\text{tr}(\mathbf{A}^{\frac{1}{2}}))^2,$$

where $\mathbf{A} = \mathbf{W}^T\mathbf{W}$. The first equality holds because of the last constraint in problem (13), and the last inequality holds because of the Cauchy-Schwarz inequality for the Frobenius norm. Moreover, $\text{tr}(\Omega^{-1}\mathbf{A})$ attains its minimum value $(\text{tr}(\mathbf{A}^{\frac{1}{2}}))^2$ if and only if $\Omega^{-\frac{1}{2}}\mathbf{A}^{\frac{1}{2}} = a\Omega^{\frac{1}{2}}$ for some constant $a$ and $\text{tr}(\Omega) = 1$. So we can get the analytical solution $\Omega = \frac{(\mathbf{W}^T\mathbf{W})^{\frac{1}{2}}}{\text{tr}((\mathbf{W}^T\mathbf{W})^{\frac{1}{2}})}$.

We set the initial value of $\Omega$ to $\frac{1}{m}\mathbf{I}_m$ which corresponds to the assumption that all tasks are unrelated initially.

After learning the optimal values of $\mathbf{W}$, $\mathbf{b}$ and $\Omega$, we can make prediction for a new data point. Given a test data point $\mathbf{x}_\star^i$ for task $T_i$, the predictive output $y_\star^i$ is given by $y_\star^i = \sum_{p=1}^{m} \sum_{q=1}^{n_p} \alpha_q^p k_{MT}(\mathbf{x}_q^p, \mathbf{x}_\star^i) + b_i$.

### 2.3 Incorporation of New Tasks

The method described above can only learn from multiple tasks simultaneously which is the setting in symmetric multi-task learning. In asymmetric multi-task learning, when a new task arrives, we could add the data for this new task to the training set and then train a new model from scratch for the $m + 1$ tasks using the above method.

However, it is undesirable to incorporate new tasks in this way due to the high computational cost incurred. Here we introduce an algorithm for asymmetric multi-task learning which is more efficient.

For notational simplicity, let $\tilde{m}$ denote $m + 1$. We denote the new task by $T_{\tilde{m}}$ and its training set by $\mathcal{D}_{\tilde{m}} = \{(\mathbf{x}_j^{\tilde{m}}, y_j^{\tilde{m}})\}_{j=1}^{n_{\tilde{m}}}$. The task covariances between $T_{\tilde{m}}$ and the $m$ existing tasks are represented by the vector $\boldsymbol{\omega}_{\tilde{m}} = (\omega_{\tilde{m},1}, \ldots, \omega_{\tilde{m},m})^T$ and the task variance for $T_{\tilde{m}}$ is defined as $\sigma$. Thus the augmented task covariance matrix for the $m + 1$ tasks is:

$$\tilde{\boldsymbol{\Omega}} = \begin{pmatrix} (1-\sigma)\boldsymbol{\Omega} & \boldsymbol{\omega}_{\tilde{m}} \\ \boldsymbol{\omega}_{\tilde{m}}^T & \sigma \end{pmatrix},$$

where $\boldsymbol{\Omega}$ is scaled by $(1 - \sigma)$ to make $\tilde{\boldsymbol{\Omega}}$ satisfy the constraint $\text{tr}(\tilde{\boldsymbol{\Omega}}) = 1$. The linear function for task $T_{m+1}$ is defined as $f_{m+1}(\mathbf{x}) = \mathbf{w}^T \mathbf{x} + b$.

With $\mathbf{W}_m = (\mathbf{w}_1, \ldots, \mathbf{w}_m)$ and $\boldsymbol{\Omega}$ at hand, the optimization problem can be formulated as follows:

$$\min_{\mathbf{w},b,\boldsymbol{\omega}_{\tilde{m}},\sigma} \quad \frac{1}{n_{\tilde{m}}} \sum_{j=1}^{n_{\tilde{m}}} l(y_j^{\tilde{m}}, \mathbf{w}^T \mathbf{x}_j^{\tilde{m}} + b) + \frac{\lambda_1}{2} \|\mathbf{w}\|_2^2$$

$$+ \frac{\lambda_2}{2} \text{tr}(\mathbf{W}_{\tilde{m}} \tilde{\boldsymbol{\Omega}}^{-1} \mathbf{W}_{\tilde{m}}^T)$$

$$\text{s.t.} \quad \tilde{\boldsymbol{\Omega}} \succeq 0, \quad (14)$$

where $\|\cdot\|_2$ denotes the 2-norm of a vector and $\mathbf{W}_{\tilde{m}} = (\mathbf{W}_m, \mathbf{w})$. Problem (14) is an SDP problem. Here we assume $\boldsymbol{\Omega}$ is positive definite.[2] So if the constraint in (14) holds, then according to the Schur complement [8], this constraint is equivalent to $\boldsymbol{\omega}_{\tilde{m}}^T \boldsymbol{\Omega}^{-1} \boldsymbol{\omega}_{\tilde{m}} \leq \sigma - \sigma^2$. Thus problem (14) becomes

$$\min_{\mathbf{w},b,\boldsymbol{\omega}_{\tilde{m}},\sigma} \quad \frac{1}{n_{\tilde{m}}} \sum_{j=1}^{n_{\tilde{m}}} l(y_j^{\tilde{m}}, \mathbf{w}^T \mathbf{x}_j^{\tilde{m}} + b) + \frac{\lambda_1}{2} \|\mathbf{w}\|_2^2$$

$$+ \frac{\lambda_2}{2} \text{tr}(\mathbf{W}_{\tilde{m}} \tilde{\boldsymbol{\Omega}}^{-1} \mathbf{W}_{\tilde{m}}^T)$$

$$\text{s.t.} \quad \boldsymbol{\omega}_{\tilde{m}}^T \boldsymbol{\Omega}^{-1} \boldsymbol{\omega}_{\tilde{m}} \leq \sigma - \sigma^2. \quad (15)$$

This is a convex problem and thus we can also use an alternating method to solve it. When using the alternating method to optimize with respect to $\mathbf{w}$ and $b$, it is similar to that in single-task learning. When optimizing with respect to $\boldsymbol{\omega}_{m+1}$ and $\sigma$, the optimization problem is formulated as

$$\min_{\boldsymbol{\omega}_{\tilde{m}},\sigma,\tilde{\boldsymbol{\Omega}}} \quad \text{tr}(\mathbf{W}_{\tilde{m}} \tilde{\boldsymbol{\Omega}}^{-1} \mathbf{W}_{\tilde{m}}^T)$$

$$\text{s.t.} \quad \boldsymbol{\omega}_{\tilde{m}}^T \boldsymbol{\Omega}^{-1} \boldsymbol{\omega}_{\tilde{m}} \leq \sigma - \sigma^2$$

$$\tilde{\boldsymbol{\Omega}} = \begin{pmatrix} (1-\sigma)\boldsymbol{\Omega} & \boldsymbol{\omega}_{\tilde{m}} \\ \boldsymbol{\omega}_{\tilde{m}}^T & \sigma \end{pmatrix}. \quad (16)$$

We impose a constraint as $\mathbf{W}_{\tilde{m}} \tilde{\boldsymbol{\Omega}}^{-1} \mathbf{W}_{\tilde{m}}^T \preceq \frac{1}{t}\mathbf{I}_d$ and the objective function becomes $\min \frac{1}{t}$ which is equivalent to $\min -t$ since $t > 0$. Using the Schur complement, we can get

$$\mathbf{W}_{\tilde{m}} \tilde{\boldsymbol{\Omega}}^{-1} \mathbf{W}_{\tilde{m}}^T \preceq \frac{1}{t}\mathbf{I}_d \iff \begin{pmatrix} \tilde{\boldsymbol{\Omega}} & \mathbf{W}_{\tilde{m}}^T \\ \mathbf{W}_{\tilde{m}} & \frac{1}{t}\mathbf{I}_d \end{pmatrix} \succeq \mathbf{0}.$$

---

[2]When $\boldsymbol{\Omega}$ is positive semi-definite, the optimization procedure is similar.

By using the Schur complement again, we get

$$\begin{pmatrix} \tilde{\boldsymbol{\Omega}} & \mathbf{W}_{\tilde{m}}^T \\ \mathbf{W}_{\tilde{m}} & \frac{1}{t}\mathbf{I}_d \end{pmatrix} \succeq \mathbf{0} \iff \tilde{\boldsymbol{\Omega}} - t\mathbf{W}_{\tilde{m}}^T \mathbf{W}_{\tilde{m}} \succeq \mathbf{0}.$$

So problem (16) can be formulated as

$$\min_{\boldsymbol{\omega}_{\tilde{m}},\sigma,\tilde{\boldsymbol{\Omega}},t} \quad -t$$

$$\text{s.t.} \quad \boldsymbol{\omega}_{\tilde{m}}^T \boldsymbol{\Omega}^{-1} \boldsymbol{\omega}_{\tilde{m}} \leq \sigma - \sigma^2$$

$$\tilde{\boldsymbol{\Omega}} = \begin{pmatrix} (1-\sigma)\boldsymbol{\Omega} & \boldsymbol{\omega}_{\tilde{m}} \\ \boldsymbol{\omega}_{\tilde{m}}^T & \sigma \end{pmatrix}.$$

$$\tilde{\boldsymbol{\Omega}} - t\mathbf{W}_{\tilde{m}}^T \mathbf{W}_{\tilde{m}} \succeq \mathbf{0}, \quad (17)$$

which is an SDP problem. In real applications, the number of tasks $m$ is usually not very large and we can use standard SDP solver to solve problem (17). Moreover we can reformulate problem (17) as a second-order cone programming (SOCP) problem [21] which is more efficient than SDP when $m$ is large. We present the procedure in the appendix.

In case two or more new tasks arrive together, the above formulation only needs to be modified slightly to accommodate all the new tasks simultaneously.

### 2.4 Kernel Extension

So far we have only considered the linear case for MTRL. In this section, we will apply the kernel trick to provide a nonlinear extension of the algorithm presented above.

The optimization problem for the kernel extension is essentially the same as problem (7), with the only difference being that the data point $\mathbf{x}_j^i$ is mapped to $\Phi(\mathbf{x}_j^i)$ in some reproducing kernel Hilbert space where $\Phi(\cdot)$ denotes the feature map. Then the corresponding kernel function $k(\cdot, \cdot)$ satisfies $k(\mathbf{x}_1, \mathbf{x}_2) = \Phi(\mathbf{x}_1)^T \Phi(\mathbf{x}_2)$.

We can also use an alternating method to solve the optimization problem. In the first step of the alternating method, we use the nonlinear multi-task kernel $k_{MT}(\mathbf{x}_{j_1}^{i_1}, \mathbf{x}_{j_2}^{i_2}) = \mathbf{e}_{i_1}^T \boldsymbol{\Omega}(\lambda_1 \boldsymbol{\Omega} + \lambda_2 \mathbf{I}_m)^{-1} \mathbf{e}_{i_2} k(\mathbf{x}_{j_1}^{i_1}, \mathbf{x}_{j_2}^{i_2})$. The rest is the same as the linear case. For the second step, the change needed is in the calculation of $\mathbf{W}^T \mathbf{W}$. Since $\mathbf{W} = \sum_{i=1}^{m} \sum_{j=1}^{n_i} \alpha_j^i \Phi(\mathbf{x}_j^i) \mathbf{e}_i^T \boldsymbol{\Omega}(\lambda_1 \boldsymbol{\Omega} + \lambda_2 \mathbf{I}_m)^{-1}$ which is similar to the representer theorem in single-task learning, we have $\mathbf{W}^T \mathbf{W} = \sum_{i,j} \sum_{p,q} \alpha_j^i \alpha_q^p k(\mathbf{x}_j^i, \mathbf{x}_q^p)(\lambda_1 \boldsymbol{\Omega} + \lambda_2 \mathbf{I}_m)^{-1} \boldsymbol{\Omega} \mathbf{e}_i \mathbf{e}_p^T \boldsymbol{\Omega}(\lambda_1 \boldsymbol{\Omega} + \lambda_2 \mathbf{I}_m)^{-1}$. When a new task arrives, the optimization problem (15) is still relevant requiring only some small changes in the calculation of $\mathbf{W}^T \mathbf{W}$. It is similar to the above calculation and so the details are omitted here.

### 2.5 Discussions

In our probabilistic model, the prior on $\mathbf{W}$ given in Eq. (2) is very general and some existing method can also be included in our model. For example, when $q(\mathbf{W}) = \mathcal{MN}_{d\times m}(\mathbf{W}|\mathbf{0}_{d\times m}, \mathbf{\Sigma}\otimes\mathbf{I}_m)$, it leads to a formulation similar to multi-task feature learning [1, 2]:

$$\min_{\mathbf{W},\mathbf{b},\mathbf{\Sigma}} \quad \sum_{i=1}^{m}\frac{1}{n_i}\sum_{j=1}^{n_i}(y_j^i - \mathbf{w}_i^T\mathbf{x}_j^i - b_i)^2 + \frac{\lambda_1}{2}\mathrm{tr}(\mathbf{W}\mathbf{W}^T)$$
$$+ \frac{\lambda_2}{2}\mathrm{tr}(\mathbf{W}^T\mathbf{\Sigma}^{-1}\mathbf{W})$$
$$\text{s.t.} \quad \mathbf{\Sigma}\succeq 0$$
$$\mathrm{tr}(\mathbf{\Sigma}) = 1.$$

From this aspect, we can understand the difference between our method and multi-task feature learning. Multi-task feature learning is to learn the covariance structure on the model parameters and the parameters of different tasks are independent given the covariance structure. However, the task relationship is not very clear in this method in that we do not know which task is helpful. In our formulation (7), the relationships between tasks are described explicitly in the task covariance matrix $\mathbf{\Omega}$. Another advantage of formulation (7) is that kernel extension is very natural as that in single-task learning. For multi-task feature learning, however, Gram-Schmidt orthogonalization on the kernel matrix is needed and hence it will incur additional computational cost.

Moreover, other matrix-variate distributions may also be used for $q(\mathbf{W})$, such as the matrix-variate $t$ distribution [14]. The resulting optimization problem is similar to problem (7) with the only difference being in the third term of the objective function. By using matrix-variate $t$ distribution, the third term becomes $\ln|\mathbf{I}_d + \mathbf{W}\mathbf{\Omega}^{-1}\mathbf{W}^T|$ making the optimization problem non-convex. Nevertheless, we can still use an alternating method to find a local optimum.

In some applications, there may exist prior knowledge about the relationships between some tasks, e.g., two tasks are more similar than two other tasks, some tasks are from the same task cluster, etc. It is easy to incorporate the prior knowledge by introducing additional constraints into problem (7). For example, if tasks $T_i$ and $T_j$ are more similar than tasks $T_p$ and $T_q$, then the corresponding constraint can be represented as $\Omega_{ij} > \Omega_{pq}$; if we know that some tasks are from the same cluster, then we can enforce the covariances between these tasks very large while their covariances with other tasks very close to 0.

## 3 Relationships with Existing Methods

Some existing multi-task learning methods [12, 11, 17, 16] also model the relationships between tasks under the regularization framework. The methods in [12, 11, 17] assume that the task relationships are given *a priori* and then utilize this prior knowledge to learn the model parameters. On the other hand, the method in [16] learns the task cluster structure from data. In this section, we discuss the relationships between MTRL and these methods.

The objective functions of the methods in [12, 11, 17, 16] are all of the following form which is similar to that of problem (7):

$$J = \sum_{i=1}^{m}\sum_{j=1}^{n_i} l(y_j^i, \mathbf{w}_i^T\mathbf{x}_j^i + b_i) + \frac{\lambda_1}{2}\mathrm{tr}(\mathbf{W}\mathbf{W}^T) + \frac{\lambda_2}{2}f(\mathbf{W}),$$

with different choices for the formulation of $f(\cdot)$.

The method in [12] assumes that all tasks are similar and so the parameter vector of each task is similar to the average parameter vector. The corresponding formulation for $f(\cdot)$ is given by

$$f(\mathbf{W}) = \sum_{i=1}^{m}\left\|\mathbf{w}_i - \frac{1}{m}\sum_{j=1}^{m}\mathbf{w}_j\right\|_2^2.$$

After some algebraic operations, we can rewrite $f(\mathbf{W})$ as

$$f(\mathbf{W}) = \sum_{i=1}^{m}\sum_{j=1}^{m}\frac{1}{2m}\|\mathbf{w}_i - \mathbf{w}_j\|_2^2 = \mathrm{tr}(\mathbf{W}\mathbf{L}\mathbf{W}^T),$$

where $\mathbf{L}$ is the Laplacian matrix defined on a fully connected graph with edge weights equal to $\frac{1}{2m}$. This corresponds to a special case of MTRL with $\mathbf{\Omega}^{-1} = \mathbf{L}$. Obviously, a limitation of this method is that only positive task correlation can be modeled.

The methods in [11] assume that the task cluster structure or the task similarity between tasks is given. $f(\cdot)$ is formulated as

$$f(\mathbf{W}) = \sum_{i,j} s_{ij}\|\mathbf{w}_i - \mathbf{w}_j\|_2^2 = \mathrm{tr}(\mathbf{W}\mathbf{L}\mathbf{W}^T),$$

where $s_{ij} \geq 0$ denotes the similarity between tasks $T_i$ and $T_j$ and $\mathbf{L}$ is the Laplacian matrix defined on the graph based on $\{s_{ij}\}$. Again, it corresponds to a special case of MTRL with $\mathbf{\Omega}^{-1} = \mathbf{L}$. Note that this method requires that $s_{ij} \geq 0$ and so it also can only model positive task correlation and task unrelatedness. If negative task correlation is modeled as well, the problem will become non-convex making it more difficult to solve. Moreover, in many real-world applications, prior knowledge about $s_{ij}$ is not available.

In [17] the authors assume the existence of a task network and that the neighbors in the task network, encoded as index pairs $(p_k, q_k)$, are very similar. $f(\cdot)$ can be formulated as

$$f(\mathbf{W}) = \sum_k \|\mathbf{w}_{p_k} - \mathbf{w}_{q_k}\|_2^2.$$

We can define a similarity matrix $G$ whose $(p_k, q_k)$th elements are equal to 1 for all $k$ and 0 otherwise. Then $f(\mathbf{W})$ can be simplified as $f(\mathbf{W}) = \mathrm{tr}(\mathbf{W}\mathbf{L}\mathbf{W}^T)$ where $\mathbf{L}$ is the Laplacian matrix of $G$, which is similar to [11]. Thus it also corresponds to a special case of MTRL with $\mathbf{\Omega}^{-1} = \mathbf{L}$. Similar to [11], a difficulty of this method is that prior knowledge in the form of a task network is not available in many applications.

The method in [16] is more general in that it learns the task cluster structure from data, making it more suitable for real-world applications. The formulation for $f(\cdot)$ is described as

$$f(\mathbf{W}) = \text{tr}\left(\mathbf{W}\left[\alpha\mathbf{H}_m + \beta(\mathbf{M} - \mathbf{H}_m) + \gamma(\mathbf{I}_m - \mathbf{M})\right]\mathbf{W}^T\right),$$

where $\mathbf{H}_m$ is the centering matrix and $\mathbf{M} = \mathbf{E}(\mathbf{E}^T\mathbf{E})\mathbf{E}^T$ with the cluster assignment matrix $\mathbf{E}$. If we let $\mathbf{\Omega}^{-1} = \alpha\mathbf{H}_m + \beta(\mathbf{M} - \mathbf{H}_m) + \gamma(\mathbf{I}_m - \mathbf{M})$ or $\mathbf{\Omega} = \frac{1}{\alpha}\mathbf{H}_m + \frac{1}{\beta}(\mathbf{M} - \mathbf{H}_m) + \frac{1}{\gamma}(\mathbf{I}_m - \mathbf{M})$, MTRL will reduce to this method. However, [16] is a local method which can only model positive task correlation within each cluster but cannot model negative task correlation. Moreover, the original optimization problem before convex relaxation is in fact non-convex and the optimal solution to the convex relaxation is not guaranteed to be the optimal solution to the original non-convex problem. Another difficulty of this method lies in determining the number of task clusters.

Compared with existing methods, MTRL is very appealing in that it can learn all three types of task relationships in a nonparametric way. This makes it easy to identify the tasks that are useful for multi-task learning and those that should not be exploited.

## 4 Experiments

In this section, we study MTRL empirically on some data sets and compare it with a single-task learning (STL) method, multi-task feature learning (MTFL) [1] method[3] and a multi-task GP (MTGP) method [7] which can also learn the global task relationships.

### 4.1 Toy Problem

We first generate a toy data set to conduct a "proof of concept" experiment before we do experiments on real data sets. The toy data set is generated as follows. The regression functions corresponding to three regression tasks are defined as $y = 3x + 10$, $y = -3x - 5$ and $y = 1$. For each task, we randomly sample five points uniformly from $[0, 10]$. Each function output is corrupted by a Gaussian noise process with zero mean and variance equal to 0.1. The data points are plotted in Figure 1, with each color (and point type) corresponding to one task. From the coefficients of the regression functions, we expect the correlation between the first two tasks to approach $-1$ and those for the other two pairs of tasks to approach $0$. To apply MTRL, we use the linear kernel and set $\lambda_1$ to 0.01 and $\lambda_2$ to 0.005. After the learning procedure converges, we find that the estimated regression functions for the three tasks are $y = 2.9964x + 10.0381$, $y = -3.0022x - 4.9421$ and $y = 0.0073x + 0.9848$. Based on the task covariance matrix learned, we obtain the following task correlation matrix:

$$\mathbf{C} = \begin{pmatrix} 1.0000 & -0.9985 & 0.0632 \\ -0.9985 & 1.0000 & -0.0623 \\ 0.0632 & -0.0623 & 1.0000 \end{pmatrix}.$$

---
[3]The implementation can be downloaded from http://www.cs.ucl.ac.uk/staff/A.Argyriou/code/.

We can see that the task correlations learned confirm our expectation, showing that MTRL can indeed learn the relationships between tasks for this toy problem.

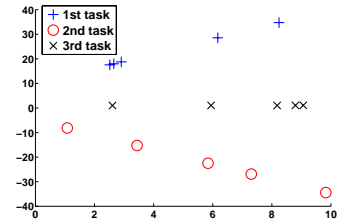

Figure 1: A toy problem. The data points with each color (and point type) correspond to one task.

### 4.2 Robot Inverse Dynamics

We now study the problem of learning the inverse dynamics of a 7-DOF SARCOS anthropomorphic robot arm[4]. Each observation in the SARCOS data set consists of 21 input features, corresponding to seven joint positions, seven joint velocities and seven joint accelerations, as well as seven joint torques for the seven degrees of freedom (DOF). Thus the input has 21 dimensions and there are seven tasks. We randomly select 600 data points for each task to form the training set and 1400 data points for each task for the test set. The performance measure used is the normalized mean squared error (nMSE), which is the mean squared error divided by the variance of the ground truth. The single-task learning method is kernel ridge regression. The kernel used is the RBF kernel. Five-fold cross validation is used to determine the values of the kernel parameter and the regularization parameters $\lambda_1$ and $\lambda_2$. We perform 10 random splits of the data and report the mean and standard derivation over the 10 trials. The results are summarized in Table 1 and the mean task correlation matrix over 10 trials is recorded in Table 2. From the results, we can see that the performance of MTRL is better than that of STL, MTFL and MTGP. From Table 2, we can see that some tasks are positively correlated (e.g., third and sixth tasks), some are negatively correlated (e.g., second and third tasks), and some are uncorrelated (e.g., first and seventh tasks).

Table 2: Mean task correlation matrix learned from SARCOS data on different tasks.

|     | 1st     | 2nd     | 3rd     | 4th     | 5th     | 6th     | 7th     |
|-----|---------|---------|---------|---------|---------|---------|---------|
| 1st | 1.0000  | 0.7435  | -0.7799 | 0.4819  | -0.5325 | -0.4981 | 0.0493  |
| 2nd | 0.7435  | 1.0000  | -0.9771 | 0.1148  | -0.0941 | -0.7772 | -0.4419 |
| 3rd | -0.7799 | -0.9771 | 1.0000  | -0.1872 | 0.1364  | 0.8145  | 0.3987  |
| 4th | 0.4819  | 0.1148  | -0.1872 | 1.0000  | -0.1889 | -0.3768 | 0.7662  |
| 5th | -0.5325 | -0.0941 | 0.1364  | -0.1889 | 1.0000  | -0.3243 | -0.2834 |
| 6th | -0.4981 | -0.7772 | 0.8145  | -0.3768 | -0.3243 | 1.0000  | 0.2282  |
| 7th | 0.0493  | -0.4419 | 0.3987  | 0.7662  | -0.2834 | 0.2282  | 1.0000  |

Moreover, we plot in Figure 2 the change in value of the objective function in problem (7). We find that the objective function value decreases rapidly and then levels off, showing the fast convergence of the algorithm which takes no more than 15 iterations.

---
[4]http://www.gaussianprocess.org/gpml/data/

Table 1: Comparison of different methods on SARCOS data. Each column represents one task. The first row of each method records the mean of nMSE over 10 trials and the second row records the standard derivation.

| Method | 1st DOF | 2nd DOF | 3rd DOF | 4th DOF | 5th DOF | 6th DOF | 7th DOF |
|---|---|---|---|---|---|---|---|
| STL | 0.2874 | 0.2356 | 0.2310 | 0.2366 | 0.0500 | 0.5208 | 0.6748 |
|  | 0.0067 | 0.0043 | 0.0068 | 0.0042 | 0.0034 | 0.0205 | 0.0048 |
| MTFL | 0.2876 | 0.1611 | 0.2125 | 0.2215 | 0.0858 | 0.5224 | 0.7135 |
|  | 0.0178 | 0.0105 | 0.0225 | 0.0151 | 0.0225 | 0.0269 | 0.0196 |
| MTGP | 0.3430 | 0.7890 | 0.5560 | 0.3147 | 0.0100 | **0.0690** | 0.6455 |
|  | 0.1038 | 0.0480 | 0.0511 | 0.1235 | 0.0067 | 0.0171 | 0.4722 |
| MTRL | **0.0968** | **0.0229** | **0.0625** | **0.0422** | **0.0045** | 0.0851 | **0.3450** |
|  | 0.0047 | 0.0023 | 0.0044 | 0.0027 | 0.0002 | 0.0095 | 0.0127 |

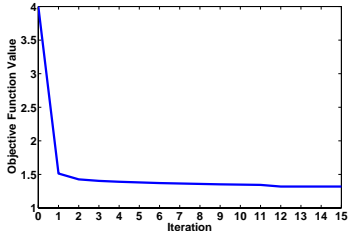

Figure 2: Convergence of objective function value for SARCOS data

### 4.3 Multi-Domain Sentiment Application

We next study a multi-domain sentiment classification application[5] which is a multi-task classification problem. Its goal is to classify the reviews of some products into two classes: positive and negative reviews. In the data set, there are four different products (tasks) from Amazon.com: books, DVDs, electronics, and kitchen appliances. For each task, there are 1000 positive and 1000 negative data points corresponding to positive and negative reviews, respectively. Each data point has 473856 feature dimensions. To see the effect of varying the training set size, we randomly select 10%, 30% and 50% of the data for each task to form the training set and the rest for the test set. The performance measure used is the classification error. We use SVM as the single-task learning method. The kernel used is the linear kernel which is widely used for text applications with high feature dimensionality. Five-fold cross validation is used to determine the values of the regularization parameters $\lambda_1$ and $\lambda_2$. We perform 10 random splits of the data and report the mean and standard derivation over the 10 trials. The results are summarized in the left column of Table 3. From the table, we can see that the performance of MTRL is better than that of STL, MTFL and MTGP on every task under different training set sizes. Moreover, the mean task correlation matrices over 10 trials for different training set sizes are recorded in the right column of Table 3. From Table 3, we can see that the first task 'books' is more correlated with the second task 'DVDs' than with the other tasks; the third and fourth tasks achieve the largest correlation among all pairs of tasks. The findings from Table 3 can be easily interpreted as follows: 'books' and 'DVDs' are mainly for entertainment; almost all the elements in 'kitchen appliances' belong to 'electronics'. So the knowledge found by our method about the relationships between tasks matches our intuition. Moreover, some interesting patterns exist in the mean task correlation matrices for different training set sizes. For example, the correlation between the third and fourth tasks is always the largest when training size varies; the correlation between the first and second tasks is larger than that between the first and third tasks, and also between the first and fourth tasks.

### 4.4 Examination Score Prediction

The school data set[6] has been widely used for studying multi-task regression. It consists of the examination scores of 15362 students from 139 secondary schools in London during the years 1985, 1986 and 1987. Thus, there are totally 139 tasks. The input consists of the year of the examination, four school-specific and three student-specific attributes. We replace each categorical attribute with one binary variable for each possible attribute value, as in [11]. As a result of this preprocessing, we have a total of 27 input attributes. The experimental settings are the same as those in [1], i.e., we use the same 10 random splits of the data to generate the training and test sets, so that 75% of the examples from each school belong to the training set and 25% to the test set. For our performance measure, we use the measure of percentage explained variance from [1], which is defined as the percentage of one minus nMSE. We use five-fold cross validation to determine the values of the kernel parameter and the regularization parameters $\lambda_1$ and $\lambda_2$. Since the experimental setting is the same, we compare our result with the results reported in [1, 7]. The results are summarized in Table 4. We can see that the performance of MTRL is better than both STL and MTFL and is slightly better than MTGP.

Table 4: Comparison of different methods on school data.

| Method | Explained Variance |
|---|---|
| STL | 23.5±1.9% |
| MTFL | 26.7±2.0% |
| MTGP | 29.2±1.6% |
| MTRL | **29.9±1.8%** |

---

[5] http://www.cs.jhu.edu/∼mdredze/datasets/sentiment/

[6] http://www.cs.ucl.ac.uk/staff/A.Argyriou/code/

Table 3: Comparison of different methods on multi-domain sentiment data for different training set sizes. The three tables in the left column record the classification errors of different methods when 10%, 30% and 50%, respectively, of the data are used for training. Each column in a table represents one task. For each method, the first row records the mean classification error over 10 trials and the second row records the standard derivation. The three tables in the right column record the mean task correlation matrices learned on different tasks for different training set sizes (10%, 30% and 50% of the data). 1st task: books; 2nd task: DVDs; 3rd task: electronics; 4th task: kitchen appliances.

| Method | 1st Task | 2nd Task | 3rd Task | 4th Task |
|---|---|---|---|---|
| STL | 0.2680 | 0.3142 | 0.2891 | 0.2401 |
|  | 0.0112 | 0.0110 | 0.0113 | 0.0154 |
| MTFL | 0.2667 | 0.3071 | 0.2880 | 0.2407 |
|  | 0.0160 | 0.0136 | 0.0193 | 0.0160 |
| MTGP | 0.2332 | 0.2739 | 0.2624 | 0.2061 |
|  | 0.0159 | 0.0231 | 0.0150 | 0.0152 |
| MTRL | **0.2233** | **0.2564** | **0.2472** | **0.2027** |
|  | 0.0055 | 0.0050 | 0.0082 | 0.0044 |

|  | 1st | 2nd | 3rd | 4th |
|---|---|---|---|---|
| 1st | 1.0000 | 0.7675 | 0.6878 | 0.6993 |
| 2nd | 0.7675 | 1.0000 | 0.6937 | 0.6805 |
| 3rd | 0.6878 | 0.6937 | 1.0000 | 0.8793 |
| 4th | 0.6993 | 0.6805 | 0.8793 | 1.0000 |

| Method | 1st Task | 2nd Task | 3rd Task | 4th Task |
|---|---|---|---|---|
| STL | 0.1946 | 0.2333 | 0.2143 | 0.1795 |
|  | 0.0102 | 0.0119 | 0.0110 | 0.0076 |
| MTFL | 0.1932 | 0.2321 | 0.2089 | 0.1821 |
|  | 0.0094 | 0.0115 | 0.0054 | 0.0078 |
| MTGP | 0.1852 | 0.2155 | 0.2088 | 0.1695 |
|  | 0.0109 | 0.0101 | 0.0120 | 0.0074 |
| MTRL | **0.1688** | **0.1987** | **0.1975** | **0.1482** |
|  | 0.0103 | 0.0120 | 0.0094 | 0.0087 |

|  | 1st | 2nd | 3rd | 4th |
|---|---|---|---|---|
| 1st | 1.0000 | 0.6275 | 0.5098 | 0.5936 |
| 2nd | 0.6275 | 1.0000 | 0.4900 | 0.5345 |
| 3rd | 0.5098 | 0.4900 | 1.0000 | 0.7286 |
| 4th | 0.5936 | 0.5345 | 0.7286 | 1.0000 |

| Method | 1st Task | 2nd Task | 3rd Task | 4th Task |
|---|---|---|---|---|
| STL | 0.1854 | 0.2162 | 0.2072 | 0.1706 |
|  | 0.0102 | 0.0147 | 0.0133 | 0.0024 |
| MTFL | 0.1821 | 0.2096 | 0.2128 | 0.1681 |
|  | 0.0095 | 0.0095 | 0.0106 | 0.0085 |
| MTGP | 0.1722 | 0.2040 | 0.1992 | 0.1496 |
|  | 0.0101 | 0.0152 | 0.0083 | 0.0051 |
| MTRL | **0.1538** | **0.1874** | **0.1796** | **0.1334** |
|  | 0.0096 | 0.0149 | 0.0084 | 0.0036 |

|  | 1st | 2nd | 3rd | 4th |
|---|---|---|---|---|
| 1st | 1.0000 | 0.6252 | 0.5075 | 0.5901 |
| 2nd | 0.6252 | 1.0000 | 0.4891 | 0.5328 |
| 3rd | 0.5075 | 0.4891 | 1.0000 | 0.7256 |
| 4th | 0.5901 | 0.5328 | 0.7256 | 1.0000 |

## 5 Conclusion

In this paper, we have presented a regularization formulation for learning the relationships between tasks in multi-task learning. Our method can model global task relationships and the learning problem can be formulated directly as a convex optimization problem. We study the proposed method in both symmetric and asymmetric multi-task learning settings.

In some multi-task learning applications, there exist additional sources of data such as unlabeled data. In our future research, we will consider incorporating additional data sources into our regularization formulation in a way similar to manifold regularization [5] to further boost the learning performance under the multi-task learning setting.

## Acknowledgments

This research has been supported by General Research Fund 622209 from the Research Grants Council of Hong Kong.

# Appendix

In this section, we show how to formulate problem (17) as a second-order cone programming (SOCP) problem.

We write $\mathbf{W}_{\tilde{m}}^T \mathbf{W}_{\tilde{m}} = \begin{pmatrix} \mathbf{\Psi}_{11} & \mathbf{\Psi}_{12} \\ \mathbf{\Psi}_{12}^T & \Psi_{22} \end{pmatrix}$ where $\mathbf{\Psi}_{11} \in \mathbb{R}^{m \times m}$, $\mathbf{\Psi}_{12} \in \mathbb{R}^{m \times 1}$ and $\Psi_{22} \in \mathbb{R}$. Then $\tilde{\mathbf{\Omega}} - t\mathbf{W}_{\tilde{m}}^T \mathbf{W}_{\tilde{m}} \succeq \mathbf{0}$ is equivalent to

$$(1-\sigma)\mathbf{\Omega} - t\mathbf{\Psi}_{11} \succeq \mathbf{0}$$
$$\sigma - t\Psi_{22} \geq (\boldsymbol{\omega}_{\tilde{m}} - t\mathbf{\Psi}_{12})^T \Big((1-\sigma)\mathbf{\Omega} - t\mathbf{\Psi}_{11}\Big)^{-1} (\boldsymbol{\omega}_{\tilde{m}} - t\mathbf{\Psi}_{12}),$$

which can be reformulated as

$$(1-\sigma)\mathbf{I}_m - t\mathbf{\Omega}^{-\frac{1}{2}}\mathbf{\Psi}_{11}\mathbf{\Omega}^{-\frac{1}{2}} \succeq \mathbf{0}$$
$$\sigma - t\Psi_{22} \geq (\boldsymbol{\omega}_{\tilde{m}} - t\mathbf{\Psi}_{12})^T \mathbf{\Omega}^{-\frac{1}{2}} ((1-\sigma)\mathbf{I}_m - t\mathbf{\Omega}^{-\frac{1}{2}}\mathbf{\Psi}_{11}\mathbf{\Omega}^{-\frac{1}{2}})^{-1}$$
$$\mathbf{\Omega}^{-\frac{1}{2}}(\boldsymbol{\omega}_{\tilde{m}} - t\mathbf{\Psi}_{12}),$$

where $\mathbf{\Omega}^{-\frac{1}{2}}$ can be computed in advance. Let $\tilde{\mathbf{\Psi}}_{11} = \mathbf{\Omega}^{-\frac{1}{2}}\mathbf{\Psi}_{11}\mathbf{\Omega}^{-\frac{1}{2}}$, $\mathbf{U}$ and $\lambda_1, \ldots, \lambda_{m-1}$ denote the eigenvector matrix and eigenvalues of $\tilde{\mathbf{\Psi}}_{11}$ with $\lambda_1 \geq \ldots \geq \lambda_{m-1} \geq 0$. Then

$$(1-\sigma)\mathbf{I}_m - t\tilde{\mathbf{\Psi}}_{11} \succeq \mathbf{0} \iff 1-\sigma \geq \lambda_1 t$$

and

$$\Big((1-\sigma)\mathbf{I}_m - t\tilde{\mathbf{\Psi}}_{11}\Big)^{-1}$$
$$= \mathbf{U}\,\mathrm{diag}\Big(\frac{1}{1-\sigma-t\lambda_1}, \ldots, \frac{1}{1-\sigma-t\lambda_{m-1}}\Big)\mathbf{U}^T.$$

Combining the above results, problem (16) is formulated as

$$\begin{aligned}
\min_{\boldsymbol{\omega}_{\tilde{m}},\sigma,\mathbf{f},t} \quad & -t \\
\text{s.t.} \quad & 1-\sigma \geq t\lambda_1 \\
& \mathbf{f} = \mathbf{U}^T \mathbf{\Omega}^{-\frac{1}{2}}(\boldsymbol{\omega}_{\tilde{m}} - t\mathbf{\Psi}_{12}) \\
& \sum_{j=1}^m \frac{f_j^2}{1-\sigma-t\lambda_j} \leq \sigma - t\Psi_{22} \\
& \boldsymbol{\omega}_{\tilde{m}}^T \mathbf{\Omega}^{-1} \boldsymbol{\omega}_{\tilde{m}} \leq \sigma - \sigma^2,
\end{aligned} \quad (18)$$

where $f_j$ is the $j$th element of $\mathbf{f}$. By introducing new variables $h_j$ and $r_j$ ($j=1,\ldots,m$), (18) is reformulated as

$$\begin{aligned}
\min_{\boldsymbol{\omega}_{\tilde{m}},\sigma,\mathbf{f},t,\mathbf{h},\mathbf{r}} \quad & -t \\
\text{s.t.} \quad & 1-\sigma \geq t\lambda_1 \\
& \mathbf{f} = \mathbf{U}^T \mathbf{\Omega}^{-\frac{1}{2}}(\boldsymbol{\omega}_{\tilde{m}} - t\mathbf{\Psi}_{12}) \\
& \sum_{j=1}^m h_j \leq \sigma - t\Psi_{22} \\
& r_j = 1-\sigma-t\lambda_j \ \forall j \\
& \frac{f_j^2}{r_j} \leq h_j \ \forall j \\
& \boldsymbol{\omega}_{\tilde{m}}^T \mathbf{\Omega}^{-1} \boldsymbol{\omega}_{\tilde{m}} \leq \sigma - \sigma^2.
\end{aligned} \quad (19)$$

Since

$$\frac{f_j^2}{r_j} \leq h_j \ (r_j, h_j > 0) \iff \left\|\begin{pmatrix} f_j \\ \frac{r_j - h_j}{2} \end{pmatrix}\right\|_2 \leq \frac{r_j + h_j}{2}$$

and

$$\boldsymbol{\omega}_{\tilde{m}}^T \mathbf{\Omega}^{-1} \boldsymbol{\omega}_{\tilde{m}} \leq \sigma - \sigma^2 \iff \left\|\begin{pmatrix} \mathbf{\Omega}^{-\frac{1}{2}}\boldsymbol{\omega}_{\tilde{m}} \\ \frac{\sigma-1}{2} \\ \sigma \end{pmatrix}\right\|_2 \leq \frac{\sigma+1}{2},$$

problem (19) is an SOCP problem [21] with $O(m)$ variables and $O(m)$ constraints. Then we can use a standard solver to solve problem (19) efficiently.